# An adaptive recursive sliding mode attitude control for tiltrotor UAV in flight mode transition based on super-twisting extended state observer

Mengshan Xie[1], Sheng Xu[1], Cheng-yue Su[1]*, Zu-yong Feng[1], Yuandian Chen[1], Zhenhua Shi[1], Jiebo Lian[1]

[1]School of physics and Optoelectronic Engineering, Guangdong University of Technology, Guangzhou, Guangdong, People's Republic of China

* E-mail: cysu@gdut.edu.cn

## Abstract

With the characteristics of vertical take-off and landing and long endurance, tiltrotor has attracted considerable attention in recent decades for its potential applications in civil and scientific research. However, the problems of strong couplings, nonlinear characteristics and mismatched disturbances inevitably exist in the tiltrotor, which bring great challenges to the controller design in transition mode. In this paper, we combined a super-twisting extended state observer (STESO) with an adaptive recursive sliding mode control (ARSMC) together to design a tiltrotor aircraft attitude system controller in transition mode using STESO-ARSMC (SAC). Firstly, the six degrees of freedom (DOF) nonlinear mathematical model of tiltrotor is established. Secondly, the states and disturbances are estimated by the STES observer. Thirdly, ARSM controller is designed to achieve finite time convergence. The Lyapunov function is used to testify the convergence of the tiltrotor UAV system. The new aspect is that the assessments of the states are incorporated into the control rules to adjust for disruptions. When compared to prior techniques, the control system proposed in this work can considerably enhance anti-disturbance performance. Finally, simulation tests are used to demonstrate the efficacy of the suggested technique.

## 1 Introduction

As a new type unmanned aerial vehicle (UAV), tiltrotor UAV is a complex multi-body system, which can change its configuration by tilting the servos to realize three different flight modes, namely hover mode, transition mode and forward mode [1,2]. Tiltrotor UAV combines the characteristics of helicopter and fixed wing aircraft, so it has the ability of vertical takeoff and landing and high-speed cruise. It has attracted considerable attention in recent decades for its potential applications in civil and scientific research [3,4].

In the references [5-10], researchers have studied the stable flight control and modeling and simulation of tiltrotor UAV in hover mode, and proved the ability of hovering and attitude control. In [11], a minimum energy controller is applied to linearize the tiltrotor model to achieve flight control in hover mode and forward mode. In [12], the intelligent position control of tilt rotor UAV is realized by using the hybrid control method of model reference adaptive controller (MRAC) and model predictive control (MPC) for hover mode and forward mode respectively. It could be seen that the modes of hover and forward have been widely studied experimentally and numerically. Transition mode is a relatively new research field, including conversion phase (from hover to forward mode) and reconversion phase (from forward to hover mode). In [13,14], the flight dynamics in transition mode is analyzed by studying the flight conditions at several discrete tilt angles. In [15-17], the effectiveness of the controller in the flight conversion phase is studied. In [2], researchers try to achieve rapid reconversion phase during tiltrotor UAV landing. According to the investigation, most authors choose one of the three flight modes of tiltrotor UAV for research, and the research on the transition process often only studies conversion phase or reconversion phase. That is because the tiltrotor UAV system has the characteristics of strong coupling, underactuated, nonlinear and so on. In practical application, there are still some problems, such as complex and uncertain model, vulnerable to external interference, actuator saturation and so on, which make the control of tiltrotor more challenging. However, in order to improve the vertical takeoff and landing capability of tiltrotor, it is also of great significance to reduce the transition time of tiltrotor from forward to hover mode, that is, to achieve fast mode transition [2]. Moreover, Yunus Govdeli's research indicated that the requirements for forward and reverse conversion phase are different [8]. And From [9], we could assume that the conversion phase and reconversion phase are two different processes. Therefore, this paper will study the attitude control of tiltrotor UAV in transition mode with both conversion phase and reconversion phase.

Due to the changes of its structure and dynamic characteristics, tiltrotor UAV has the characteristics of high nonlinearity, time-varying flight dynamics and inertia / control coupling, especially in the transition mode. Therefore, an effective controller is the key to complete the transition task [18]. Several control approaches have been applied on the tiltrotor aircraft flight control system, some of which include: Higher Harmonic Control [18], Model Predictive Control [19], adaptive control techniques [20], Linear Quadratic Regulator and Linear Quadratic Gaussian (LQG) approaches [4,5], classical Proportional-Integral-Derivative (PID) controller [21], Neural and Fuzzy Control [13], minimum energy controllers [22], sliding mode control (SMC) [6,10], etc. However, the major approaches listed above require an accurate mathematical model, which is difficult to achieve for a variety of reasons. As a result, an effective control strategy must be devised.

In [23], a nonlinear extended state observer (ESO) proposed by Han has been widely used in many applications for its characteristics of simple structure and high estimation efficiency, which can estimate the total disturbances based on output data. It could be seen that the observer could obtain good effects in some simulation results. In [24-27], an adaptive sliding mode observer (ASMO), a second-order sliding mode observer (SOSMO) and a time-varying discrete sliding mode observer (TVDSMO) are employed to reduce the chattering. In [28], a strict Lyapunov super twisting algorithm (SLSTA) based observer has been proposed. In [29], an adaptive super-twisting disturbance observer (ASTDO) is designed to estimate the system disturbances in fixed time, and the adaptive method relaxes the assumption about the system disturbances. In [30], a super-twisting ESO (STESO) is designed to estimate total disturbances such that estimated errors of disturbances converge to zero in finite time.

SMC has been widely used in many motion control systems because of its ability of fast convergence and strong robustness against disturbance. Nevertheless, the chattering phenomenon of the system is the main disadvantage of the general SMC [31]. Another problem of the general SMC is its relatively long asymptotic convergence property. This problem can be avoided by using the recursive control structure, in which the reaching phase is eliminated while ensuring the finite-time convergence [32]. In [33], the applications of terminal sliding mode (TSM) control have been extended to a variety of systems like ground and flight vehicles. However, the main drawback of the TSM approaches is the singularity problem of the controller that limits its implementation. To solve this problem, Khawwaf et al. [34] employed a nonsingular terminal sliding mode (NTSM) controller for the tracking control of IPMC actuators. Most of the methods mentioned, the signum function explicitly exists in the discontinuous control law that may degrade the control signal smoothness. In [35], an adaptive full-order recursive terminal sliding-mode (AFORTSM) controller is designed to develop a fast-response, high-precision, and chattering-free sliding mode control scheme.

Based on the existing research, we combined a super-twisting extended state observer (STESO) with an adaptive recursive sliding mode control (ARSMC) together to design a tiltrotor aircraft attitude system controller using SAC.

To begin, a tiltrotor UAV nonlinear model of motion is developed, taking into account model uncertainties and unknown disturbances. Second, for the tiltrotor attitude control system, a SAC control method is presented. Furthermore, the Lyapunov function is used to testify the convergence of the tiltrotor UAV system, and the sufficient conditions for the tracking error to approach zero are obtained. Finally, simulation tests are carried out to validate the strong robustness, fast convergence, as well as superior error tracking of the designed SAC.

The main contribution of this paper are as follows:

1) Different from ref [1], which only control conversion phase, or ref [19], which only control reconversion phase, or ref [20], which only discuss full envelop flying without illustrate the transition mode in detail, the proposed SAC is first used in tiltrotor UAV for the transition mode attitude control with both conversion phase and reconversion phase.

2) The proposed strategy adopts the cascade method of STESO and ARSMC, in which STESO is used to estimate the states and total disturbances of the system, so that the disturbance errors converge to zero in a finite time, and its outputs are also used to replace the states requirements in ARSMC. And the strategy has been proved by the stability of Lyapunov function. According to the authors' best of knowledge, there is no such method in the literature that attempts to realize tiltrotor UAV attitude control in the whole transition mode.

3) Compared with the fast terminal sliding mode control (FTSMC) of a quadrotor UAV [36] and the recursive sliding mode control (RSMC) of a linear motor [37], the SAC is proposed in this work for the tiltrotor UAV to reduce the convergence time of attitude in a faster manner, displaying excellent robustness against model uncertainties and unknown disturbances.

4) The effectiveness of the proposed control strategy is verified by the comparative simulation demonstrations, making the proposed SAC practically applicable for the tiltrotor UAV, or even other complex nonlinear systems.

This paper is organized as follows. In section 2, the nonlinear dynamic model of the tiltrotor UAV is introduced. In section 3, the design of STESO- ARSMC is discussed. In section 4, the stability analysis of the proposed ARSMC is elaborated. In section 5, the comparative simulation demonstrations are given. Finally, conclusions are drawn in Section 6.

## 2 Dynamics Modeling

### 2.1 Model Description

In this paper, a compact electrically powered tiltrotor UAV is presented in Figure 1. The tiltrotor specified in this paper contains six actuators, including four rotors and two tilting servos, and uses a traditional v-tail fixed-wing layout. To make the airplane physically sound, the rotors are attached to the main structural sections. The front two rotors can be tilted from 0 to 90 degrees by tilting the servos, allowing the mode transition between the hover and forward mode to occur. Illustrated in Figure 2, the tiltrotor UAV cruises in the forward mode, takes off and lands vertically in the hover mode, and enters the conversion and reconversion stages in the transition mode. The tiltrotor is controlled by four rotors and two tilting servos without the use of control surfaces in hover mode, while in the forward mode, the tiltrotor is controlled by aerodynamic control surfaces: v-tail and aileron. In the transition mode, the tiltrotor is controlled by rotors, tilting servos and aerodynamic control surfaces. We employed system identification and direct measurement method to obtain the parameters and aerodynamic coefficients of the aircraft.

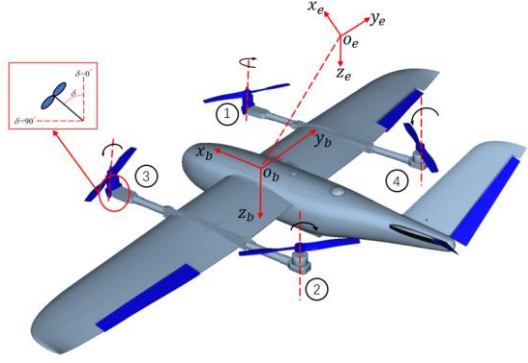

Figure 1. Tiltrotor UAV prototype view.

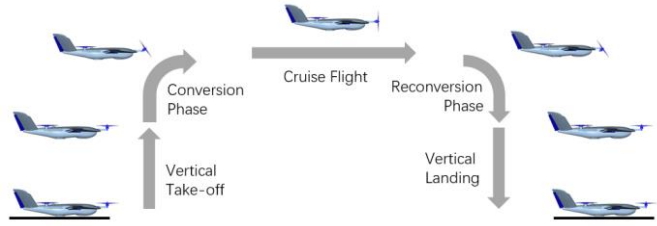

Figure 2. Tiltrotor UAV flight phase.

## 2.2 Nonlinear Equations of Motion

Several assumptions are made to effectively portray the dynamic characteristics of the tiltrotor UAV while avoiding excessive complexity [8].

**Assumption 1.** The tiltrotor UAV is a rigid body, ignoring the elastic deformation of the body;

**Assumption 2.** The aerodynamic interference between rotor and wing and between left and right rotors are not considered;

**Assumption 3.** The tilt rotor UAV is symmetrical along the body axis, and the inertia products $I_{xy}, I_{xz}$ and $I_{yz}$ are smaller than those of $X_b$, $Y_b$ and $Z_b$, so $I_{xy} = I_{xz} = I_{yz} = 0$.

This section focuses on developing the tiltrotor UAV's 6-DOF nonlinear mathematical model. Figure 1 depicts a schematic diagram of the tiltrotor coordinate system. The world frame is $O_e X_e Y_e Z_e$, the body frame is $O_b X_b Y_b Z_b$, and the rotor frame is $O_r X_r Y_r Z_r$. It is noted that the rotors are labeled by 1, 2, 3, and 4, respectively.

The transformation matrices among the coordination frames are given as Eq. (1):

$$R_e^b = \begin{bmatrix} c\theta c\psi & c\theta s\psi & -s\theta \\ s\theta c\psi s\phi - s\psi c\phi & s\theta s\psi s\phi + c\psi c\phi & c\theta s\phi \\ s\theta c\psi c\phi + s\psi s\phi & s\theta s\psi c\phi - c\psi s\phi & c\theta c\phi \end{bmatrix},$$

$$R_{ri}^b = \begin{cases} \begin{bmatrix} 1 & 0 & 0 \\ 0 & 1 & 0 \\ 0 & 0 & 1 \end{bmatrix} & (i = 2,4) \\ \begin{bmatrix} c\delta & 0 & -s\delta \\ 0 & 1 & 0 \\ s\delta & 0 & c\delta \end{bmatrix} & (i = 1,3) \end{cases} \quad (1)$$

where $\phi, \theta,$ and $\psi$ represent the Euler angles, $\delta$ is the tilting angle of the tilting steering gear, $c$ and $s$ are the symbols of cosine function and sine function, respectively. The dynamic equation in the body coordinate system can be described as Eq. (2) [38]:

$$\begin{cases}
\begin{bmatrix} \dot{u} \\ \dot{v} \\ \dot{w} \end{bmatrix} = \begin{bmatrix} rv - qw \\ pw - ru \\ qu - pv \end{bmatrix} + \frac{1}{m}\begin{bmatrix} F_x \\ F_y \\ F_z \end{bmatrix} \\
\begin{bmatrix} \dot{p} \\ \dot{q} \\ \dot{r} \end{bmatrix} = \begin{bmatrix} \frac{I_y - I_z}{I_x}qr + \frac{1}{I_x}\tau_x \\ \frac{I_z - I_x}{I_y}pr + \frac{1}{I_y}\tau_y \\ \frac{I_x - I_y}{I_z}pq + \frac{1}{I_z}\tau_z \end{bmatrix} \\
\begin{bmatrix} \dot{\phi} \\ \dot{\theta} \\ \dot{\psi} \end{bmatrix} = \begin{bmatrix} 1 & s\phi t\theta & c\phi t\theta \\ 0 & c\phi & -s\phi \\ 0 & s\phi/c\theta & c\phi/c\theta \end{bmatrix}\begin{bmatrix} p \\ q \\ r \end{bmatrix} \\
\begin{bmatrix} \dot{P}_n \\ \dot{P}_E \\ \dot{H} \end{bmatrix} = \begin{bmatrix} c\theta c\psi & s\phi s\theta c\psi - c\phi\psi & s\phi s\psi + c\phi s\theta c\psi \\ c\theta s\psi & s\phi s\theta\psi + c\phi c\psi & -s\phi c\psi + c\phi s\theta s\psi \\ s\theta & -s\phi c\theta & -c\phi c\theta \end{bmatrix}\begin{bmatrix} u \\ v \\ w \end{bmatrix}
\end{cases} \quad (2)$$

where $F = [F_x, F_y, F_z]^T$ and $M = [\tau_x, \tau_y, \tau_z]^T$ represent the vectors of total force and moment in the body frame, respectively, and $m$ is the mass. $[P_n, P_E, H]^T$ is the position vector in the world frame, $[u, v, w]^T$ and $[p, q, r]^T$ are the speed and angular velocity vectors in the body frame, respectively. $t$ is the symbol of tangent function.

$F$ in Eq. (2) can be expressed as:
$$F = F_g + F_p + F_a \quad (3)$$
where $F_g$ is the gravity, $F_p$ is the rotor thrust, and $F_a = [-D, Y, -L]^T$ is the aerodynamic force. The expressions of them are from Eq. (4) to Eq. (6).

$$F_g = R_e^b G = \begin{bmatrix} -s\theta \\ c\theta s\phi \\ c\theta c\phi \end{bmatrix} mg \quad (4)$$

$$F_p = \sum_{i=1}^{4} R_{ri}^b T_i^a = \begin{bmatrix} T_1 s\delta + T_3 s\delta \\ 0 \\ -T_1 c\delta - T_2 - T_3 c\delta - T_4 \end{bmatrix} \quad (5)$$

$$F_a = \begin{bmatrix} -D \\ Y \\ -L \end{bmatrix} = \bar{q}S\begin{bmatrix} -C_x \\ C_y \\ -C_z \end{bmatrix} \quad (6)$$

where $T_i = k_t \Omega_i^2 (i = 1,2,3,4)$ are the thrust values, $k_t$ is the rotor force coefficient, $\Omega_i (i = 1,2,3,4)$ are rotation speeds, $\bar{q} = \frac{1}{2}\rho V^2$ is the dynamic pressure, $\rho$ is the air density, $V$ is the airspeed, $S$ is the wing area, $C_x$, $C_y$, and $C_z$ are aerodynamic force coefficients along $X_b$, $Y_b$ and $Z_b$, respectively.

$M$ in Eq. (2) could be expressed as:
$$M = M_p + M_a \quad (7)$$
where $M_p$ denotes the torque vector generated by the propulsion systems, $M_a$ is created by the aerodynamic forces of the wings. the expressions of them are from Eq. (8) to Eq. (10).

$$M_p = M_{pt} + M_{pd}$$

$$M_{pt} = \sum_{i=1}^{4}(d_i \times R_{ri}^b T_i^a) = \begin{bmatrix} -dT_1 c\delta + dT_2 + dT_3 c\delta - dT_4 \\ dT_1 c\delta - dT_2 + dT_3 c\delta - dT_4 \\ -dT_1 s\delta + dT_3 s\delta \end{bmatrix} \quad (8)$$

$$M_{pd} = \sum_{i=1}^{4} R_{ri}^{b}\tau_i^a = \begin{bmatrix} \tau_1 s\delta - \tau_3 s\delta \\ 0 \\ -\tau_1 c\delta - \tau_2 + \tau_3 c\delta + \tau_4 \end{bmatrix} \quad (9)$$

$$M_a = \bar{q}S \begin{bmatrix} C_l \\ C_m \\ C_n \end{bmatrix} \quad (10)$$

where $M_{pt}$ is the torque vector caused by the thrust, $d_i$ are position vectors from the center of gravity to the axis of each rotor, $\tau_i = k_d \Omega_i^2 (i = 1,2,3,4)$ indicate the moment of force generated by rotor air resistance, $k_d$ is the rotor torque coefficient, $M_{pd}$ is the moment vector in the body frame. $C_l$, $C_m$, and $C_n$ are aerodynamic moment coefficients of roll, pitch, yaw, respectively.

## 3. Control Design

In this study, the tiltrotor attitude controller is separated into 3 parts. First, a STESO is used to estimate the states and errors. Second, an ARSMC is developed to realize robust control under the disturbances. Third, a control allocation scheme for mapping virtual control commands to aircraft control inputs is proposed. Figure 3 depicts a block schematic of the designed control method.

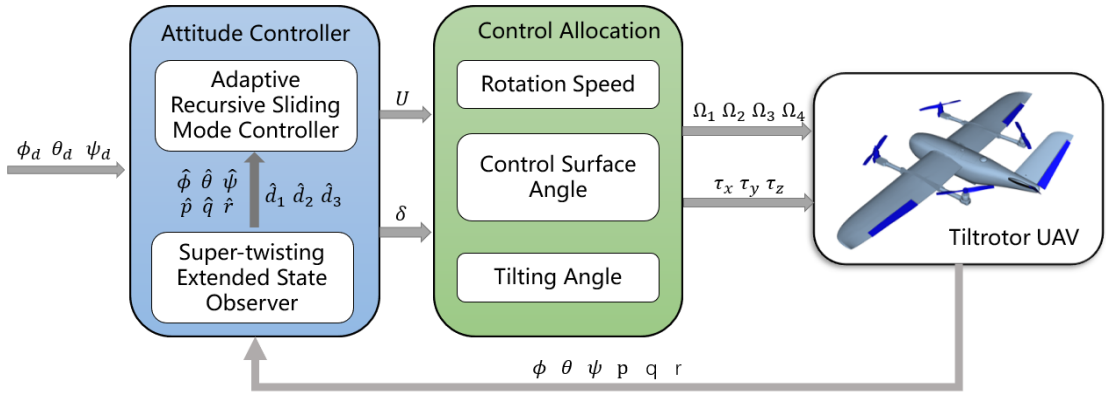

Figure 3. Structure of the overall control scheme.

**Assumption 4 [39].** The pitch angle and roll angle are quite modest and change around zero. At this point, the tiltrotor dynamic model can be simplified as a second-order nonlinear system as $[\ddot{\phi}, \ddot{\theta}, \ddot{\psi}]^T = [\dot{p}, \dot{q}, \dot{r}]^T$.

According to Assumption 4, the tiltrotor UAV attitude model of roll channel can be described in the state-space form as

$$\begin{cases} \dot{\phi} = p \\ \dot{p} = \frac{I_y - I_z}{I_x} qr + \frac{1}{I_x} u_1 + d_1 \\ \dot{\theta} = q \\ \dot{q} = \frac{I_z - I_x}{I_y} pr + \frac{1}{I_y} u_2 + d_2 \\ \dot{\psi} = r \\ \dot{r} = \frac{I_x - I_y}{I_z} pq + \frac{1}{I_z} u_3 + d_3 \end{cases} \quad (11)$$

**Assumption 5.** Assuming that the lumped disturbances $d_i (i = 1,2,3)$ are bounded by a known positive constant $c$ such that $|d_i| \leq c$, and $|\dot{d}_i| \leq \Delta$.

## 3.1 Super-Twisting Extended State Observer (STESO)

In this section, we take the roll channel as an example for controller design, the other two channels including pitch channel and yaw channel are similar. Considering the nonlinear system in Eq. (11), whose states $\phi$, $p$ and lumped disturbances $d_1$ can be estimated through a STESO given by [40]

$$\begin{cases} \dot{\hat{\phi}} = \hat{p} - h_1 \text{sig}^{\frac{2}{3}}(\tilde{\phi}) \\ \dot{\hat{p}} = \hat{d}_1 + \frac{I_y - I_z}{I_x} qr + \frac{1}{I_x} u_1 - h_2 \text{sig}^{\frac{1}{3}}(\tilde{\phi}) \\ \dot{\hat{d}}_1 = -h_3 \text{sign}(\tilde{\phi}) \end{cases} \tag{12}$$

where $\tilde{\phi} = \phi - \hat{\phi}$. Considering the estimation error dynamics for system in Eq. (12), which can be written as

$$\begin{cases} \dot{\tilde{\phi}} = \tilde{p} - h_1 \text{sig}^{\frac{2}{3}}(\tilde{\phi}) \\ \dot{\tilde{p}} = \tilde{d}_1 - h_2 \text{sig}^{\frac{1}{3}}(\tilde{\phi}) \\ \dot{\tilde{d}}_1 = -h_3 \text{sign}(\tilde{\phi}) + \dot{d}_1 \end{cases} \tag{13}$$

where $\tilde{p} = p - \hat{p}$ and $\tilde{d}_1 = d_1 - \hat{d}_1$.

The above equation is finite-time stable, which has been proved in [41]. Thus, when the appropriate gains $h_1, h_2,$ and $h_3$ being selected, $\tilde{\phi}$, $\tilde{p}$, and $\tilde{d}_1$ will converge to zero in finite time $t > T_0$. The estimation errors eventually converge to zero in finite time.

## 3.2 Adaptive Recursive Sliding Mode Control (ARSMC)

Since $\tilde{d}_i = d_i - \hat{d}_i (i = 1,2,3)$, Eq. (11) can be rewrite as

$$\begin{cases} \dot{\phi} = p \\ \dot{p} = \frac{I_y - I_z}{I_x} qr + \frac{1}{I_x} u_1 + \tilde{d}_1 + \hat{d}_1 \\ \dot{\theta} = q \\ \dot{q} = \frac{I_z - I_x}{I_y} pr + \frac{1}{I_y} u_2 + \tilde{d}_2 + \hat{d}_2 \\ \dot{\psi} = r \\ \dot{r} = \frac{I_x - I_y}{I_z} pq + \frac{1}{I_z} u_3 + \tilde{d}_3 + \hat{d}_3 \end{cases} \tag{14}$$

To construct the ARSM controller, which is built by the recursive terminal sliding surfaces, as follows [13, 26]:

$$\begin{cases} s_1 = \sigma_1 + \lambda_1 \eta_1 \\ s_2 = \sigma_2 + \lambda_2 \eta_2 \\ s_3 = \sigma_3 + \lambda_3 \eta_3 \end{cases} \tag{15}$$

with fast terminal sliding functions $\sigma_i (i = 1,2,3)$ given by

$$\begin{cases} \sigma_1 = \dot{e}_\phi + k_1 e_\phi + k_2 \int sig^{\alpha_1}(e_\phi) \\ \sigma_2 = \dot{e}_\theta + k_3 e_\theta + k_4 \int sig^{\alpha_2}(e_\theta) \\ \sigma_3 = \dot{e}_\psi + k_5 e_\psi + k_6 \int sig^{\alpha_3}(e_\psi) \end{cases} \tag{16}$$

$$\begin{cases} \dot{\eta}_1 = sig^{\beta_1}(\sigma_1) \\ \dot{\eta}_2 = sig^{\beta_2}(\sigma_2) \\ \dot{\eta}_3 = sig^{\beta_3}(\sigma_3) \end{cases} \tag{17}$$

where $k_i (i = 1, \ldots, 6)$ are positive constants, $\alpha_i (i = 1,2,3) > 1$, $\lambda_i (i = 1,2,3) > 0$, $0 < \beta_i (i = 1,2,3) < 1$, $e_\phi = \phi - \phi_d$, $e_\theta = \theta - \theta_d$, and $e_\psi = \psi - \psi_d$. The initial values of $\eta_i (i = 1,2,3)$ are designed as:

$$\begin{cases} \eta_1(0) = -\lambda_1^{-1}\sigma_1(0) \\ \eta_2(0) = -\lambda_2^{-1}\sigma_2(0) \\ \eta_3(0) = -\lambda_3^{-1}\sigma_3(0) \end{cases} \quad (18)$$

**Remark 1.** By substituting Eq. (18) into Eq. (16), the initial sliding variables $s_i(0) = 0 (i = 1,2,3)$ can be verified, which means that the control system is forced to start on the sliding surfaces at the initial time, so that to eliminate the arrival time [37].

The overall control inputs of the RTSMC can be built as

$$\begin{cases} u_1 = u_{01} + u_{11} \\ u_2 = u_{02} + u_{12} \\ u_3 = u_{03} + u_{13} \end{cases} \quad (19)$$

Sensors can readily get the attitude data in real applications. As a result, the starting values of $\eta_i (i = 1,2,3)$ may be computed by Eq. (15) and Eq. (18).

$$\begin{cases} \eta_1(0) = -\lambda_1^{-1}[\dot{e}_\phi(0) + k_1 e_\phi(0) + k_2 \int sig^{\alpha_1}(e_\phi(0))] \\ \eta_2(0) = -\lambda_2^{-1}[\dot{e}_\theta(0) + k_3 e_\theta(0) + k_4 \int sig^{\alpha_2}(e_\theta(0))] \\ \eta_3(0) = -\lambda_3^{-1}[\dot{e}_\psi(0) + k_5 e_\psi(0) + k_6 \int sig^{\alpha_3}(e_\psi(0))] \end{cases} \quad (20)$$

Considering Eq. (14), the derivative of sliding surfaces in Eq. (15) becomes:

$$\begin{cases} \dot{s}_1 = \dot{\sigma}_1 + \lambda_1 \dot{\eta}_1 \\ \quad = \ddot{e}_\phi + k_1 e_\phi + k_2 \int sig^{\alpha_1}(e_\phi) + \lambda_1 \dot{\eta}_1 \\ \dot{s}_2 = \dot{\sigma}_2 + \lambda_2 \dot{\eta}_2 \\ \quad = \ddot{e}_\theta + k_3 e_\theta + k_4 \int sig^{\alpha_2}(e_\theta) + \lambda_2 \dot{\eta}_2 \\ \dot{s}_3 = \dot{\sigma}_3 + \lambda_3 \dot{\eta}_3 \\ \quad = \ddot{e}_\psi + k_5 e_\psi + k_6 \int sig^{\alpha_3}(e_\psi) + \lambda_2 \dot{\eta}_2 \end{cases} \quad (21)$$

Considering that the equivalent controls lead to $\dot{s}_i (i = 1,2,3) = 0$, the following expressions can be generated from Eq. (21)

$$\begin{cases} \dot{\sigma}_1 = -\lambda_1 \dot{\eta}_1 \\ \dot{\sigma}_2 = -\lambda_2 \dot{\eta}_2 \\ \dot{\sigma}_3 = -\lambda_3 \dot{\eta}_3 \end{cases} \quad (22)$$

By setting $\tilde{d}_i (i = 1,2,3) = 0$, we may get the following equivalent control inputs

$$\begin{cases} u_{01} = -(I_y - I_z)qr + I_x \ddot{\phi}_d - I_x \left[ k_1 \dot{e}_\phi + k_2 sig^{\alpha_1}(e_\phi) + \hat{d}_1 + \int \lambda_1^2 \beta_1 |\sigma_1|^{2\beta_1 - 1} \right] \\ u_{02} = -(I_z - I_x)pr + I_y \ddot{\theta}_d - I_y \left[ k_3 \dot{e}_\theta + k_4 sig^{\alpha_2}(e_\theta) + \hat{d}_2 + \int \lambda_2^2 \beta_2 |\sigma_2|^{2\beta_2 - 1} \right] \\ u_{03} = -(I_x - I_y)pq + I_z \ddot{\psi}_d - I_z \left[ k_5 \dot{e}_\psi + k_6 sig^{\alpha_3}(e_\psi) + \hat{d}_3 + \int \lambda_3^2 \beta_3 |\sigma_3|^{2\beta_3 - 1} \right] \end{cases} \quad (23)$$

However, the system states will be far away from the sliding surfaces even though $s_i (i = 1,2,3) = 0$ for the existence of disturbances. The switching control laws are selected as Eq. (24) to increase the ability of disturbance rejection.

$$\begin{cases} u_{11} = -I_x \left[ \hat{\xi}_1 s_1 + \hat{\xi}_2 \int sig^{\frac{1}{2}}(\dot{s}_1) \right] \\ u_{12} = -I_y \left[ \hat{\xi}_3 s_2 + \hat{\xi}_4 \int sig^{\frac{1}{2}}(\dot{s}_2) \right] \\ u_{13} = -I_z \left[ \hat{\xi}_5 s_3 + \hat{\xi}_6 \int sig^{\frac{1}{2}}(\dot{s}_3) \right] \end{cases} \quad (24)$$

where $\hat{\xi}_i (i = 1,2,3,4,5,6)$ are updated by the following adaptive laws.

$$\begin{cases} \dot{\hat{\xi}}_1 = -\varrho \dot{s}_1^2 \\ \dot{\hat{\xi}}_2 = -\varrho sig^{\frac{1}{2}}(\dot{s}_1) \\ \dot{\hat{\xi}}_3 = -\varrho \dot{s}_2^2 \\ \dot{\hat{\xi}}_4 = -\varrho sig^{\frac{1}{2}}(\dot{s}_2) \\ \dot{\hat{\xi}}_5 = -\varrho \dot{s}_3^2 \\ \dot{\hat{\xi}}_6 = -\varrho sig^{\frac{1}{2}}(\dot{s}_3) \end{cases} \tag{25}$$

where $\varrho$ is positive constant. The estimation errors of $\hat{\xi}_i (i = 1,2,3,4,5,6)$ with respect to the desired values $\xi_{id}(i = 1,2,3,4,5,6)$ are constants and defined as

$$\tilde{\xi}_i = \xi_{id} - \hat{\xi}_i \ (i = 1,2,3,4,5,6) \tag{26}$$

where it is assumed that the conditions

$$\xi_{1d} \geq \frac{|\dot{d}_1|}{|\dot{s}_1|}, \ \xi_{2d} > 0, \ \xi_{3d} \geq \frac{|\dot{d}_3|}{|\dot{s}_3|}, \ \xi_{4d} > 0, \ \xi_{5d} \geq \frac{|\dot{d}_5|}{|\dot{s}_5|}, \ \xi_{6d} > 0 \tag{27}$$

are met.

## 3.3 Control Allocation

The control allocation technique is presented in this section to map from the control commands to the tiltrotor UAV's manipulated inputs. In forward mode, the control allocation follows the standard fixed wing allocation approach, thus we just need to pay attention to the hover mode. $T, u_1, u_2,$ and $u_3$ are the virtual control instructions for the tiltrotor UAV, where $T$ is directly connected to altitude control while $u_1, u_2,$ and $u_3$ are related to roll control, pitch control and yaw control, respectively. Six actuators, comprising four rotors and two servos, may be utilized to control the flying of tiltrotor UAV. The computation of real outputs can be separated into two sections due to the different response speeds of the rotors and the servos. $T, u_1, u_2,$ and $u_3$ acquire the rotation speeds of four rotors during the first section. As shown in Figure 1, the four rotors are labeled by 1, 2, 3 and 4, respectively. It should be noted that the tilting angles are estimated before calculating the rotation speed $[\Omega_1, \Omega_2, \Omega_3, \Omega_4]^T$. By

$$\begin{bmatrix} u_1 \\ u_2 \\ u_3 \\ T \end{bmatrix} = R \begin{bmatrix} \Omega_1 \\ \Omega_2 \\ \Omega_3 \\ \Omega_4 \end{bmatrix} \tag{28}$$

where

$$R = \begin{bmatrix} -dk_t c\delta - k_d s\delta & dk_t & dk_t c\delta - k_d s\delta & -dk_t \\ dk_t c\delta & dk_t & -dk_t c\delta & -dk_t \\ -dk_t s\delta - k_d c\delta & -k_d & dk_t s\delta + k_d c\delta & k_d \\ -k_t c\delta & -k_t & -k_t c\delta & -k_t \end{bmatrix} \tag{29}$$

Because $R$ is a square matrix, the essential point in determining the rotation speed $[\Omega_1, \Omega_2, \Omega_3, \Omega_4]^T$ is to ensure that $R$ is reversible. The determinant of $R$ can be written as

$$|R| = 8dk_t^2[d^2k_t^2 s\delta c\delta + dk_t k_d (c\delta)^2] \tag{30}$$

In hover and transition modes, the tilting angle is limited to $(0°, 90°)$, allowing the matrix to be deduced as $|R| \neq 0$. Hence the rotation speed may be calculated by

$$\begin{bmatrix} \Omega_1 \\ \Omega_2 \\ \Omega_3 \\ \Omega_4 \end{bmatrix} = R^{-1} \begin{bmatrix} u_1 \\ u_2 \\ u_3 \\ T \end{bmatrix} \tag{31}$$

For the second part, the tilting angle needs to be determined based on airspeed $V$, which is usually designed according to the experiments. In this study, the scheduling rule is presented as

$$\ddot{\delta}(t) = \begin{cases} a, & 0 \leqslant t - t_0 < \frac{b}{a} \\ 0, & \frac{b}{a} \leqslant t - t_0 < -\frac{\pi}{2b} \\ -a, & -\frac{\pi}{2b} \leqslant t - t_0 < -\frac{\pi}{2b} + \frac{b}{a} \end{cases}$$

$$\dot{\delta}(t) = \int_{t_0}^{t} \ddot{\delta}(\tau) d\tau$$

$$\delta(t) = \int_{t_0}^{t} \dot{\delta}(\tau) d\tau + \begin{cases} \frac{\pi}{2}, & conversion \\ 0, & reconversion \end{cases}$$

(32)

where $a$ and $b$ are both constants, and $a < 0$, $b < 0$.

## 4 Stability Analysis

The conclusion of the presented control method is stated in the following theorem, which also includes a stability analysis.

**Lemma 1 [42].** Considering a system, the positive Lyapunov function meets the condition:

$$\dot{V}(t) + k(V(t))^\zeta \leq 0 \quad \forall t \leq t_0, V(t_0) \geq 0 \tag{32}$$

where $0 < \zeta < 1, k > 0$, the system is finite-time stable. Then, for any given $t_0$, $V(t)$ meets

$$V^{1-\zeta}(t) \leq V^{1-\zeta}(t_0) - k(1-\zeta)(t-t_0), \quad t_0 \leq t \leq t_1 \tag{33}$$

And $V(t) \equiv 0$ for any given $t \geq t_V$, where the convergence time $t_V$ of $V(t)$ from the initial value $V(0)$ to zero is

$$t_V = t_0 + \frac{V^{(1-\zeta)}(t_0)}{k(1-\zeta)} \tag{34}$$

**Theorem 1.** Considering the tiltrotor UAV flight system in Eq. (11) and overall control inputs of the RTSMC in Eq. (19), the state tracking error $e_\phi, e_\theta, e_\psi$ converge to zero in a finite time.

In this section, we take the roll channel as an example for stability analysis, the other two channels including pitch channel and yaw channel are similar.

**Proof.** Define the Lyapunov functions as follows:

$$V_1 = \frac{1}{2}\dot{s}_1^2 + \frac{1}{2}\tilde{\xi}_1^2 + \frac{1}{2}\tilde{\xi}_2^2 \tag{35}$$

The derivative of Eq. (35) may be computed as

$$\dot{V}_1 = \dot{s}_1\ddot{s}_1 + \tilde{\xi}_1\dot{\tilde{\xi}}_1 + \tilde{\xi}_2\dot{\tilde{\xi}}_3$$
$$= \dot{s}_1\ddot{s}_1 + \tilde{\xi}_1\dot{\tilde{\xi}}_1 + \tilde{\xi}_2\dot{\tilde{\xi}}_2 \tag{36}$$

Taking the first-order and second-order time derivatives of $\sigma_1$ in Eq. (16) yields

$$\dot{\sigma}_1 = \ddot{e}_\phi + k_1\dot{e}_\phi + k_2 sig^{\alpha_1}(e_\phi) \tag{37}$$

$$\ddot{\sigma}_1 = \dddot{e}_\phi + k_1\ddot{e}_\phi + k_2\alpha_1|e_\phi|^{\alpha_1-1}\dot{e}_\phi \tag{38}$$

In view of Eq. (14), the first-order, second-order and third-order time derivatives of $e_\phi$ can be obtained

$$\dot{e}_\phi = \dot{\phi} - \dot{\phi}_d \tag{39}$$

$$\ddot{e}_\phi = \frac{I_y - I_z}{I_x}qr + \frac{1}{I_x}u_1 + \tilde{d}_1 + \hat{d}_1 - \ddot{\phi}_d \tag{40}$$

$$\dddot{e}_\phi = \frac{d}{dt}(\frac{I_y-I_z}{I_x}qr) + \frac{1}{I_x}\dot{u}_1 + \dot{\tilde{d}}_1 + \dot{\hat{d}}_1 - \dddot{\phi}_d \tag{41}$$

Substituting Eq. (41) into Eq. (38) leads to

$$\ddot{\sigma}_1 = \frac{d}{dt}(\frac{I_y - I_z}{I_x}qr) + \frac{1}{I_x}\dot{u}_1 + \dot{d}_1 + \dot{d}_1 - \dddot{\phi}_d \\ + k_1\ddot{e}_\phi + k_2\alpha_1|e_\phi|^{\alpha_1 - 1}\dot{e}_\phi \tag{42}$$

The second-order time derivative of $s$ in Eq. (15) is

$$\ddot{s}_1 = \ddot{\sigma}_1 + \lambda_1\ddot{\eta}_1 \tag{43}$$

Then, substituting Eq. (22), Eq. (42) and the first-order time derivative of the control signal u in Eq. (19) into Eq. (43) yields

$$\ddot{s}_1 = -(\hat{\xi}_1\dot{s}_1 + \hat{\xi}_2 sig^{\frac{1}{2}}(\dot{s}_1) - \dot{\tilde{d}}_1) \tag{44}$$

Then, substituting Eq. (25), and Eq. (44) into Eq. (36) leads to

$$\dot{V}_1 = -\dot{s}_1(\hat{\xi}_1\dot{s}_1 + \hat{\xi}_2 sig^{\frac{1}{2}}(\dot{s}_1) - \dot{\tilde{d}}_1) + \tilde{\xi}_1\dot{\hat{\xi}}_1 + \tilde{\xi}_2\dot{\hat{\xi}}_2 \\ = -\dot{s}_1\left(\hat{\xi}_1\dot{s}_1 + \hat{\xi}_2 sig^{\frac{1}{2}}(\dot{s}_1) - \dot{\tilde{d}}_1\right) + \tilde{\xi}_1(-\varrho\dot{s}_1^2) + \tilde{\xi}_2\left(-\varrho sig^{\frac{1}{2}}(\dot{s}_1)\right) \tag{45}$$

Taking $\varrho = 1$, and considering Eq. (26) yields

$$\dot{V}_1 = -\xi_{1d}\dot{s}_1^2 - \xi_{2d}sig^{\frac{1}{2}}(\dot{s}_1) + \dot{\tilde{d}}_1\dot{s}_1 \tag{46}$$

According to Eq. (27)

$$\dot{V}_1 = -\xi_{2d}sig^{\frac{1}{2}}(\dot{s}_1) - \dot{s}_1\left(\xi_{1d}\dot{s}_1 - \dot{\tilde{d}}_1\right) \leq -\xi_{2d}sig^{\frac{1}{2}}(\dot{s}_1) \leq 0 \tag{47}$$

After the adaptation period $t_a$, $\hat{\xi}_1$ and $\hat{\xi}_2$ achieve their intended values $\xi_{1d}$ and $\xi_{2d}$, respectively, and the estimate errors $\tilde{\xi}_1$ and $\tilde{\xi}_2$ disappear. Then the Eq. (35) becomes

$$V_1 = \frac{1}{2}\dot{s}_1^2 \tag{48}$$

Substituting Eq. (48) into Eq. (47) leads to

$$\dot{V}_1 + 2^{\frac{1}{4}}\xi_{2d}V_1^{\frac{1}{4}} \leq 0 \tag{49}$$

As a result, Eq. (49) meets the criterion of finite time convergence in Lemma 1. Define $t_{V_1}$ as the time for $V_1$ to converge to zero, including the time for $s_1$, $\tilde{\xi}_1$ and $\tilde{\xi}_2$ to converge to zero, respectively, given as.

$$t_{V_1} \leq \frac{V_1^{(1-\frac{1}{4})}(0)}{2^{\frac{1}{4}}\xi_{2d}(1-\frac{1}{4})} \tag{50}$$

Furthermore, when $s_1(0) = 0$, according to Eq. (15), we can obtain $\sigma_1 + \lambda_1\eta_1 = 0$. By using Eq. (17), we have

$$\dot{\sigma}_1 + \lambda_1|\sigma_1|^{\beta_1}\text{sign}(\sigma_1) = 0 \tag{51}$$

According to Lemma 1, the time taken for $\sigma$ converging to the zero is $t_{s_1}$,

$$t_{s_1} \leq \frac{|\sigma_1(0)|^{1-\beta_1}}{\lambda_1(1-\beta_1)} \tag{52}$$

Once $\sigma_1 = 0$, from Eq. (16), it is obvious that $\dot{e}_\phi = -k_1 e_\phi - k_2 \int sig^{\alpha_1}(e_\phi)$. Define the Lyapunov function $V_{11} = \frac{1}{2}e_\phi^2$. The derivative of $V_{11}$ is calculated as

$$\dot{V}_{11} = e_\phi\dot{e}_\phi = -k_1 e_\phi^2 - k_2 e_\phi \int sig^{\alpha_1}(e_\phi) \leq -\sqrt{2}(k_1|e_\phi| + k_2 \int sig^{\alpha_1}(e_\phi))\frac{|e_\phi|}{\sqrt{2}} = -\Lambda V_{11}^{\frac{1}{2}} \tag{53}$$

where $\Lambda = \sqrt{2}(k_1|e_\phi| + k_2 \int sig^{\alpha_1}(e_\phi)) > 0$. According to Lemma 2, the time for $e_\phi$ converging to the zero is $t_{\sigma_1}$

$$t_{\sigma_1} \leq \frac{2V_{11}^{\frac{1}{2}}(0)}{\Lambda} \tag{54}$$

At the end, the entire time when the tracking error of $\phi$ converge to zero within the time $t_\phi = t_{V_1} + t_{s_1} + t_{\sigma_1}$.

The proof of Theorem 1 is complete.

**Remark 2.** To decrease chattering, function $sat(\cdot)$ is used instead of $sign(\cdot)$ in Eq. (24), which is written as:

$$sat(x) = \begin{cases} 1, & x > \Delta \\ \frac{1}{\Delta}x, & |x| \leq \Delta \\ -1, & x < \Delta \end{cases} \tag{55}$$

**Remark 3.** From the adaptive rule in Eq. (25), it could be seen that the adaptive time $t_a$ depends on the control parameter $\varrho$. Usually, the larger the parameter $\varrho$, the faster the adaptation process will be. In fact, due to sensor noises, an infinite adaptive procedure may occur, resulting in excessively huge gains $\hat{\xi}_1$ and $\hat{\xi}_2$. In order to overcome this issue, the adaptive rule is enhanced in practice as follows:

$$\begin{aligned} \dot{\hat{\xi}}_1 &= \begin{cases} -\varrho \dot{s}_1^2, & if \ |e_\phi| > \varepsilon \\ 0, & if \ |e_\phi| \leq \varepsilon \end{cases} \\ \dot{\hat{\xi}}_2 &= \begin{cases} -\varrho sig^{\frac{1}{2}}(\dot{s}_1), & if \ |e_\phi| > \varepsilon \\ 0, & if \ |e_\phi| \leq \varepsilon \end{cases} \end{aligned} \tag{56}$$

where $\varepsilon$ is the specified threshold of the tracking error.

**Remark 4.** The recursive structure proposed in this paper also aids in reducing chattering in the control input. The PID sliding function in Eq. (16) is selected, as indicated in Eq. (24).

Unlike the conventional switching element, the integral operation of the switching element is used in the arrival control input in Eq. (24), and this characteristic improves the smoothness of the control input signal even further.

## 5 Simulation Tests

### 5.1 Tests scenario

The model parameters of the tiltrotor UAV are specified as shown in Table 1.

Table 1. the tiltrotor UAV model parameters.

| Parameter | Definition | Value |
|---|---|---|
| $m$ | Mass | $6 \ kg$ |
| $g$ | Gravity | $9.81 \ m/s^2$ |
| $l$ | Wing span | $2.1 \ m$ |
| $S$ | Wing aera | $0.48 \ m^2$ |
| $\bar{c}$ | Chord length | $0.25 \ m$ |
| $I_x$ | Inertia around $x$ axis | $0.876 \ kg.m^2$ |
| $I_y$ | Inertia around $y$ axis | $0.166 \ kg.m^2$ |
| $I_z$ | Inertia around $z$ axis | $0.115 \ kg.m^2$ |

In the simulation, the tilting process begins at 2 seconds and ends at 8 seconds. During the tilting process, the forces and moments always change.

1) The UAV initialization settings

Suppose the initial speeds are $[u, v, w]^T = [0, 0, 0]^T\ m/s$, the initial flight angular velocities are $[p, q, r]^T = [0, 0, 0]^T\ deg/s$, the initial flight Euler angles are $[\phi, \theta, \psi]^T = [0.57, 0.57, 1.14]^T\ deg$, and $a = -4, b = -14$ in the conversion phase, while $a = -180, b = -90$ in the reconversion phase. The external disturbance is given in Eq. (57). Shown in Figure 4, the simulation lasts 24 seconds which contains the whole process of the transition. The flight phases are divided by four dashed lines at 3, 13, and 20, 22s in the simulation results and the same below, which mean 3-13s is the conversion phase and 20-22s is the reconversion phase.

$$d = \begin{cases} 5\sin(\pi(t-9)), & 9 \leq t \leq 11 \\ 0, & else \end{cases} \quad (57)$$

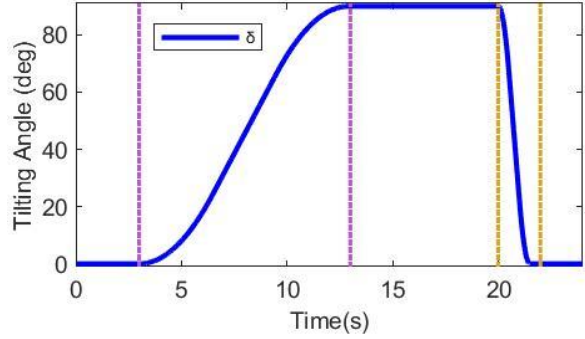

Figure 4. Tiltrotor UAV transition mode.

2) The gains of the proposed SAC

Tables 2 and 3 show the proposed STES observer and ARSM controller settings, respectively.

Table 2. STESO parameters.

| Parameter | Value |
| --- | --- |
| $\delta$ | 0.01 |
| $\alpha_1$ | 0.5 |
| $\alpha_2$ | 0.25 |
| $\beta_1$ | 30 |
| $\beta_2$ | 300 |
| $\beta_3$ | 1000 |

Table 3. ARSMC parameters.

| Parameter | Value |
| --- | --- |
| A | 71 |
| c | 1.2 |
| d | 0.5 |
| e | -0.1 |
| k | -5 |

3) The simulation conditions

The desired flight angular velocities are $[p, q, r]^T = [0, 0, 0]^T\ deg/s$, and the desired flight Euler angle are $[\phi, \theta, \psi]^T = [0, 0, 0]^T\ deg$.

4) The quantitative evaluation

To quantitatively evaluate the tracking control performance of the proposed SAC, two common performance indicators: the maximum absolute tracking error ($MAX_e$) and the root mean square ($RMS_e$) are introduced, which are described by the following equations:

$$MAX_e = \max_{k=1,\dots,N}(|e(k)|) \tag{58}$$

$$RMS_e = \sqrt{\frac{1}{N}\sum_{k=1}^{N}e^2(k)} \tag{59}$$

where $k$ is the sample index, and $N$ is the number of samples. We also compare the performance of the proposed SAC to that of the FTSMC and RSMC, which were previously described in Ref. [36] and Ref. [37], respectively.

## 5.2 Simulation Tests

In this part, we use the simulation of MATLAB R2019a / Simulink to verify the effectiveness and robustness of the designed control scheme SAC during the tilting process of the tiltrotor UAV. In addition, robustness tests of the proposed SAC are carried out in this section.

1) Attitude tracking tests without disturbance.

The test results of attitude tracking without disturbance during the tilting process are as shown in Figure 5 to 10.

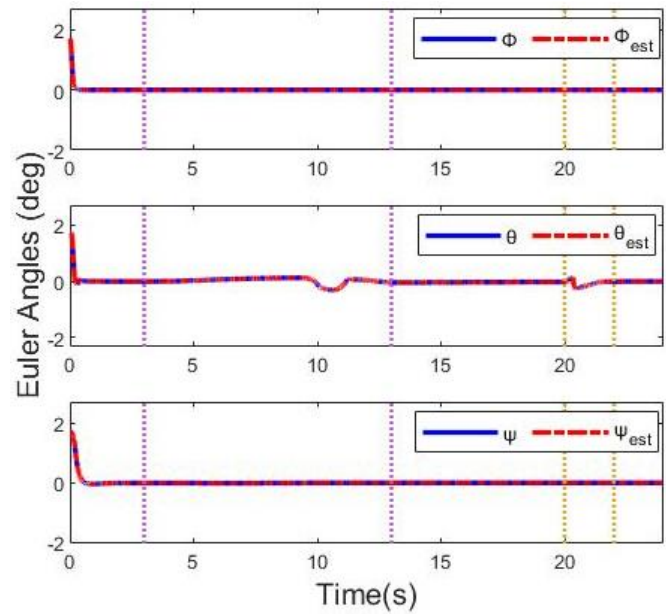

Figure 5. Euler angle estimate curves without disturbance

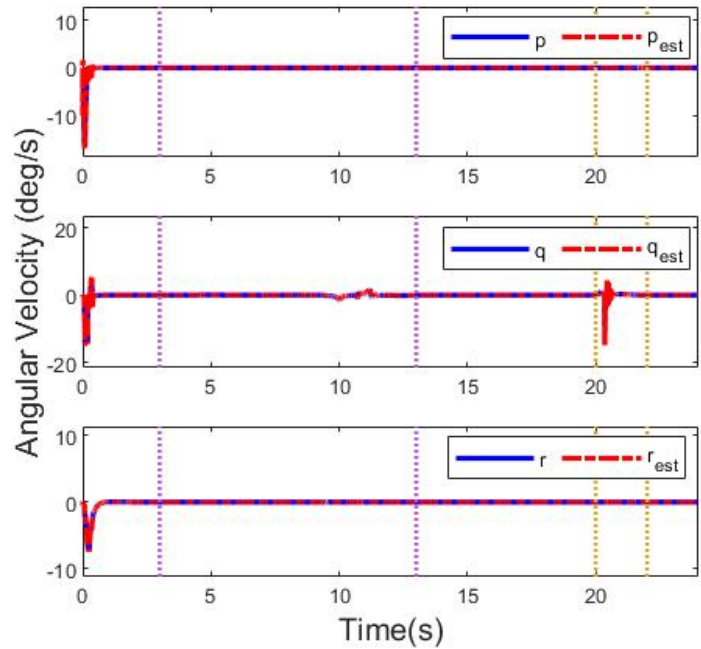

Figure 6. Angular velocity estimate curves without disturbance

Figure 5 and 6 illustrate the comparison curve between the actual states and estimate states of Euler angles and angular velocities, respectively. And the STES observer accurately estimates the states of the tiltrotor UAV during the transition mode. Moreover, it could be seen that the transition mode has a great impact on the control of pitch channel, which is due to the conflict between the control laws of forward mode and hover mode during the transition process, resulting in a pitch of 0.3 degrees.

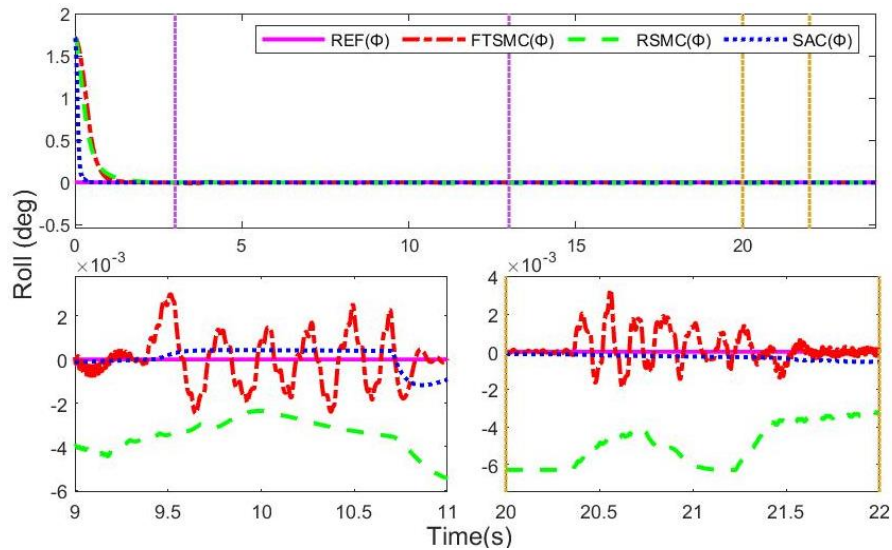

Figure 7. Comparison of three control methods without disturbance in roll channel, including two detail plots of 9-11s and 20-22s

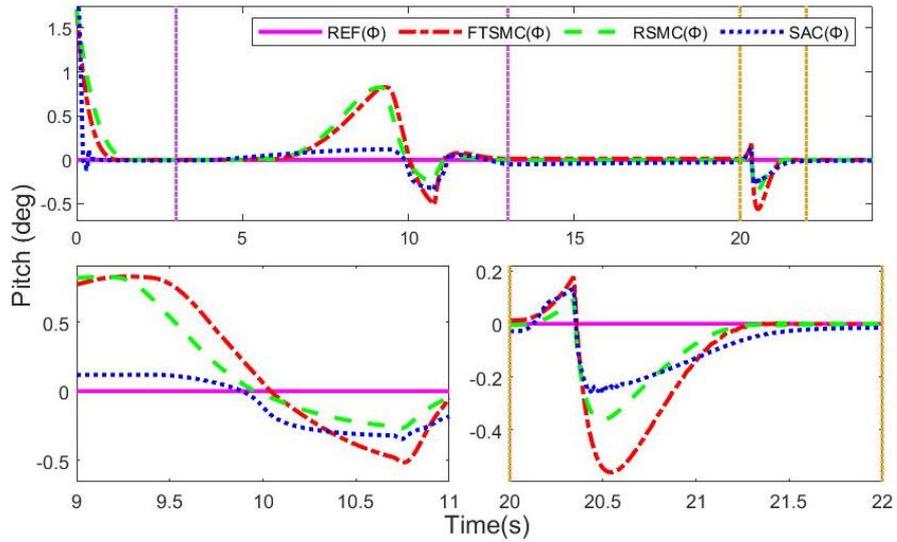

Figure 8. Comparison of three control methods without disturbance in pitch channel, including two detail plots of 9-11s and 20-22s

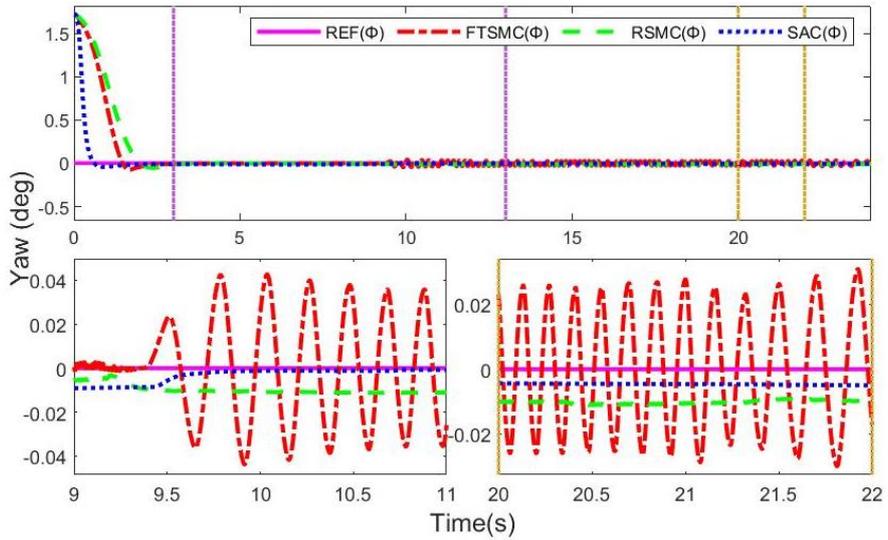

Figure 9. Comparison of three control methods without disturbance in yaw channel, including two detail plots of 9-11s and 20-22s

Shown from Figure 7 to 9, the FTSMC, the RSMC and the proposed SAC can effectively track the roll angle, pitch angle and yaw angle. Among that, from the details of three channels, the proposed controller has the best performance in the transition mode. On the one hand, the proposed SAC has the fastest convergence time and converges to the expected value at about 0.7 second, while FTSMC and RSMC converge at about 1.9 and 2.5 seconds respectively. On the other hand, SAC has the best robustness and the chatter of pitch angle is relatively small.

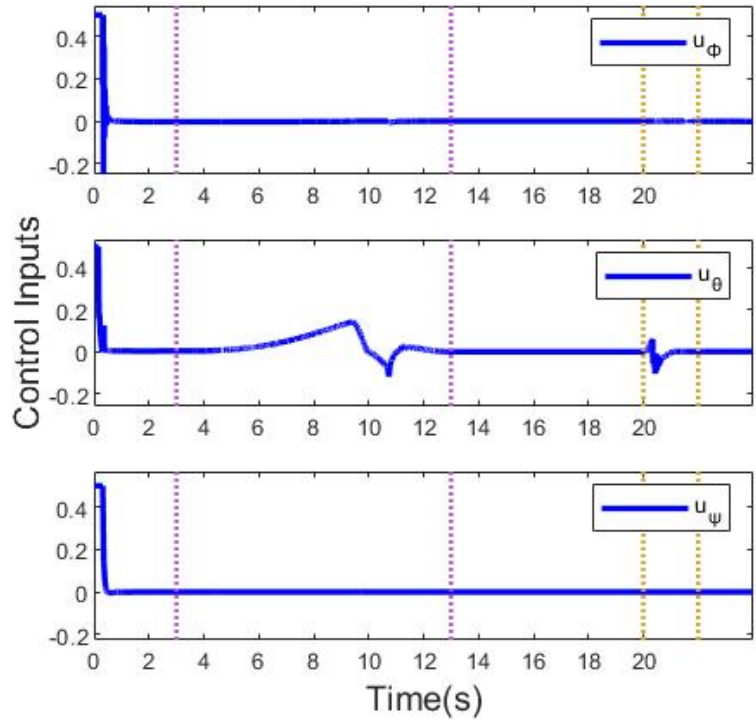

Figure 10. Developed control laws without disturbance

Figure 10 shows the control inputs for tracking the reference attitudes, which are smooth, continuous, and easy to implement in practice.

2) Attitude tracking tests with disturbance.

The test results of attitude tracking with disturbance during the transition process are as shown in Figure 11 to 16.

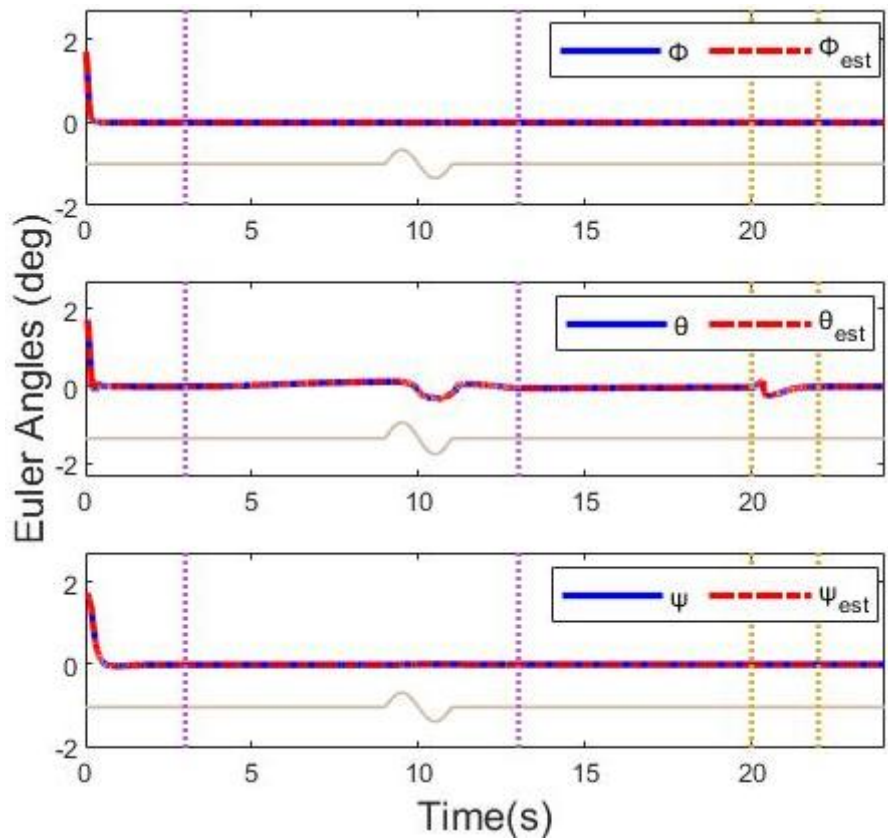

Figure 11. Euler angle estimate curves with disturbance

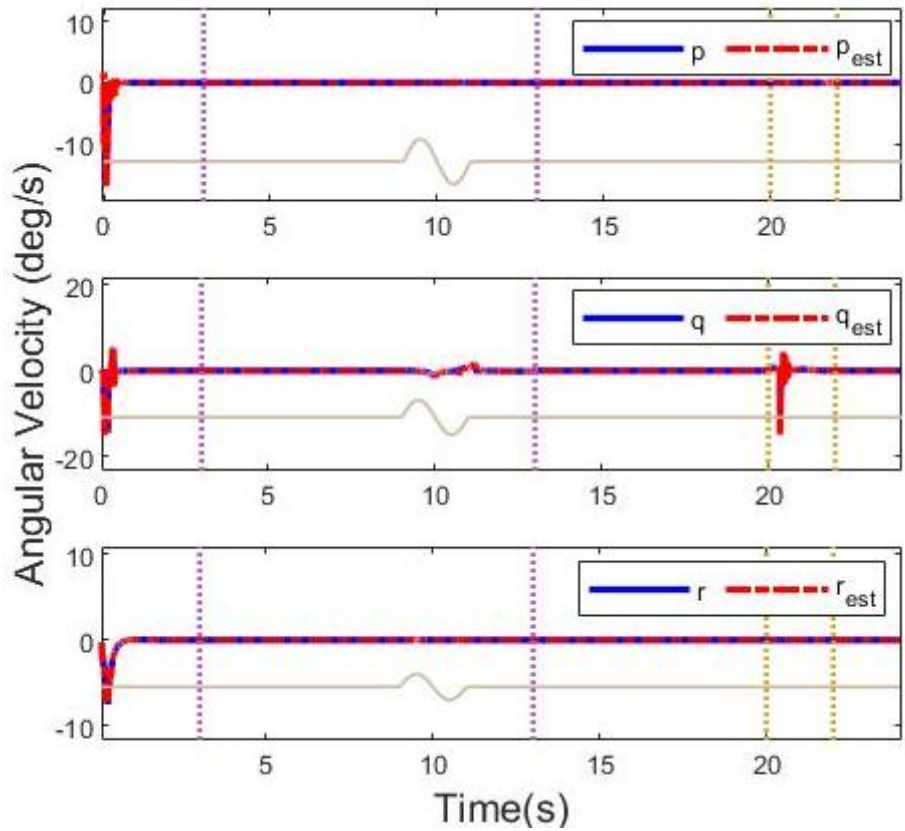

Figure 12. Angular velocity estimate curves with disturbance

Compared with Figure 5 and 6, Figure 9 and 10 describe the Euler angle and angular velocity estimation curves under disturbance. Based on the observer, the proposed SAC presents excellent robustness against model uncertainties and external disturbances.

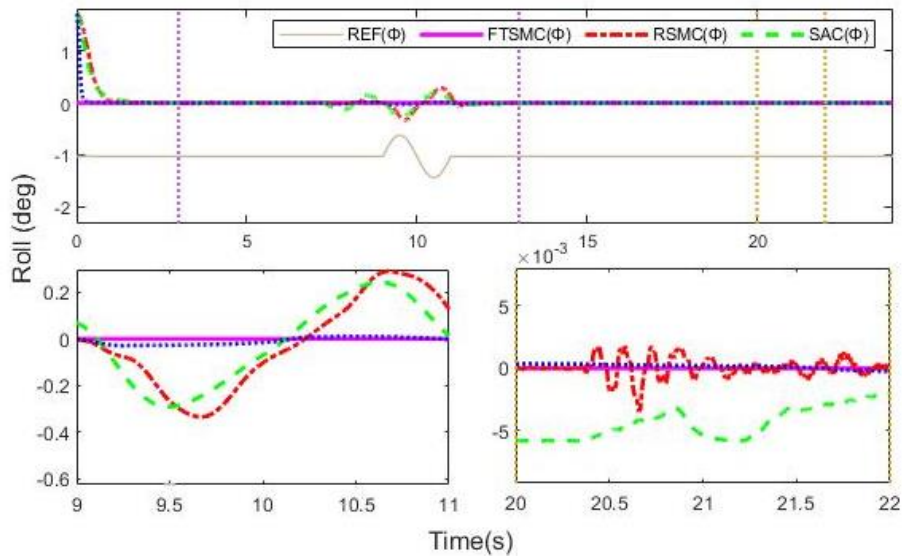

Figure 13. Comparison of three control methods with disturbance in roll channel, including two detail plots of 9-11s and 20-22s

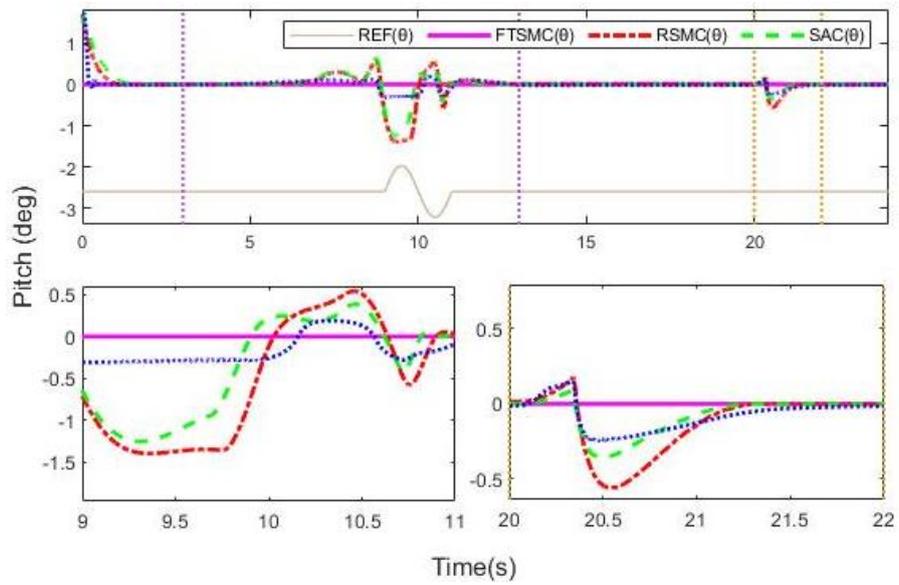

Figure 14. Comparison of three control methods with disturbance in roll channel, including two detail plots of 9-11s and 20-22s

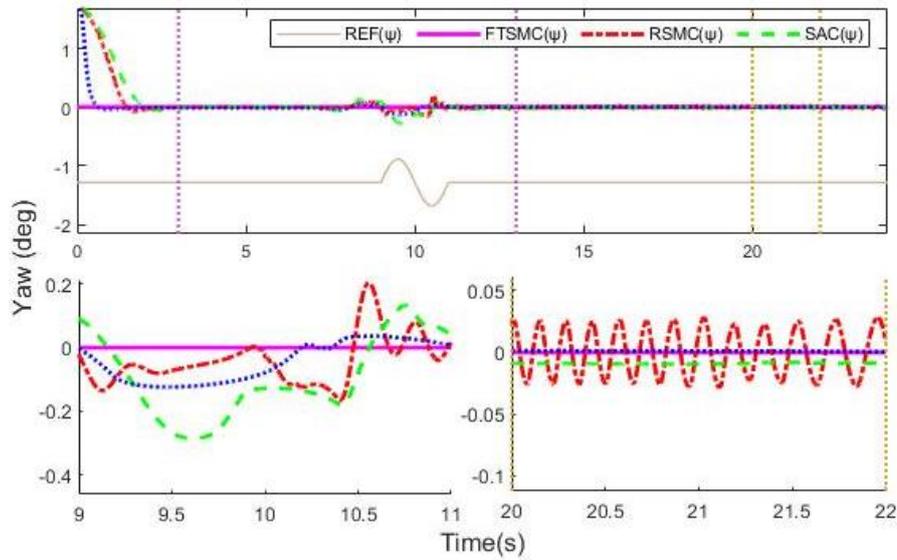

Figure 15. Comparison of three control methods with disturbance in roll channel, including two detail plots of 9-11s and 20-22s

Compared with Figure 7 to 9, Figure 13 to 15 show that the FTSMC, the RSMC and the proposed SAC can still effectively track attitudes of tiltrotor UAV under the condition of disturbance, among that SAC shows better performance, more accurate attitude tracking and stronger robustness. And the detail plots in the three channels show that the proposed SAC converges to a small neighborhood of the desire value in finite time. Compared with the other two control methods, the proposed controller has the best anti-interference performance.

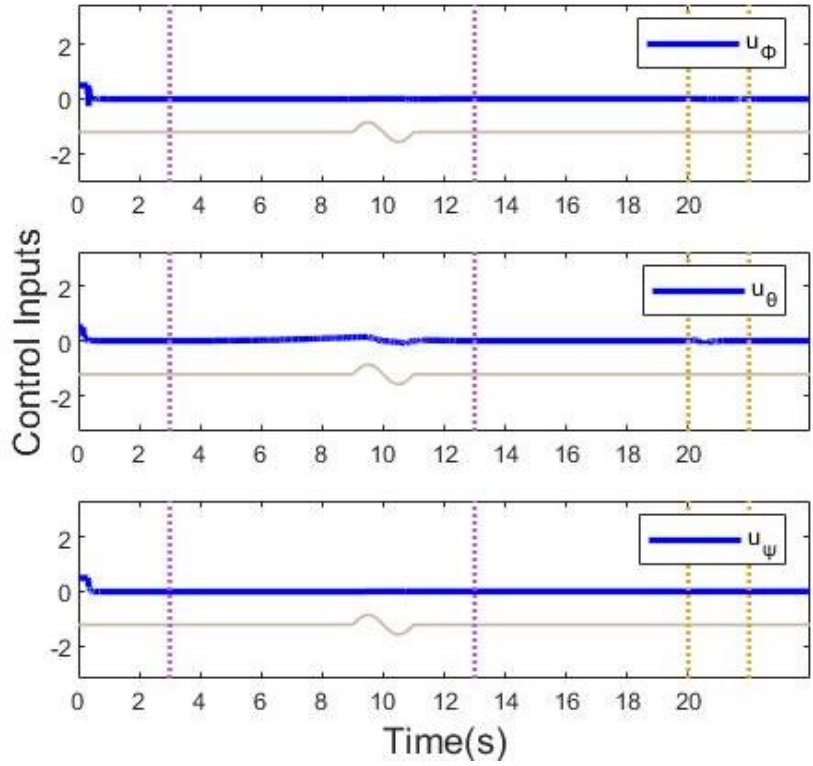

Figure 16. Developed control laws with disturbance.

Compared with Figure. 10, Figure. 16 shows that even in the case of interference, the control inputs of the proposed controller are still smooth, continuous, and easy to implement.

Table 4. Comparison of test results in conversion phase without disturbance.

| Channel | | FTSMC | RSMC | SAC | I1 | I2 |
|---|---|---|---|---|---|---|
| Roll | $MAX_e$ | 0.011 | 0.006 | 0.001 | 0.89 | 0.81 |
| | $RMS_e$ | 0.002 | 0.004 | 0.000 | 0.84 | 0.91 |
| Pitch | $MAX_e$ | 0.828 | 0.824 | 0.345 | 0.58 | 0.58 |
| | $RMS_e$ | 0.286 | 0.279 | 0.102 | 0.64 | 0.63 |
| Yaw | $MAX_e$ | 0.044 | 0.014 | 0.013 | 0.70 | 0.70 |
| | $RMS_e$ | 0.015 | 0.008 | 0.008 | 0.47 | 0.00 |

Table 5. Comparison of test results in conversion phase with disturbance.

| Channel | | FTSMC | RSMC | SAC | I1 | I2 |
|---|---|---|---|---|---|---|
| Roll | $MAX_e$ | 0.335 | 0.291 | 0.029 | 0.91 | 0.90 |
| | $RMS_e$ | 0.081 | 0.084 | 0.009 | 0.88 | 0.89 |
| Pitch | $MAX_e$ | 1.396 | 1.252 | 0.310 | 0.78 | 0.75 |
| | $RMS_e$ | 0.384 | 0.312 | 0.116 | 0.70 | 0.63 |
| Yaw | $MAX_e$ | 0.203 | 0.286 | 0.124 | 0.39 | 0.57 |
| | $RMS_e$ | 0.042 | 0.070 | 0.035 | 0.17 | 0.51 |

Table 6. Comparison of test results in reconversion phase without disturbance.

| Channel | | FTSMC | RSMC | SAC | I1 | I2 |
|---|---|---|---|---|---|---|
| Roll | $MAX_e$ | 0.003 | 0.006 | 0.001 | 0.71 | 0.85 |
|  | $RMS_e$ | 0.001 | 0.004 | 0.001 | 0.05 | 0.85 |
| Pitch | $MAX_e$ | 0.561 | 0.363 | 0.256 | 0.54 | 0.29 |
|  | $RMS_e$ | 0.165 | 0.102 | 0.088 | 0.47 | 0.13 |
| Yaw | $MAX_e$ | 0.032 | 0.011 | 0.006 | 0.83 | 0.49 |
|  | $RMS_e$ | 0.020 | 0.010 | 0.005 | 0.75 | 0.50 |

Table 7. Comparison of test results in reconversion phase with disturbance.

| Channel |  | FTSMC | RSMC | SAC | I1 | I2 |
|---|---|---|---|---|---|---|
| Roll | $MAX_e$ | 0.003 | 0.006 | 0.001 | 0.75 | 0.85 |
|  | $RMS_e$ | 0.001 | 0.003 | 0.000 | 0.19 | 0.87 |
| Pitch | $MAX_e$ | 0.561 | 0.360 | 0.252 | 0.55 | 0.30 |
|  | $RMS_e$ | 0.166 | 0.101 | 0.088 | 0.47 | 0.13 |
| Yaw | $MAX_e$ | 0.032 | 0.010 | 0.001 | 0.96 | 0.86 |
|  | $RMS_e$ | 0.019 | 0.009 | 0.001 | 0.96 | 0.91 |

The performance of the three controllers in this paper is summarized in Table 4 to Table 7, where I1 and I2 represent the improvement compare with FTSMC and RSMC, respectively. According to these four tables, compared with FTSMC and RSMC, the suggested SAC achieves superior performance (namely, the $MAX_e$ and $RMS_e$ of the SAC are the smallest) in tiltrotor UAV attitude control of transition mode with and without disturbance. Compared with FTSMC, the improvement ratios of SAC are ranging from 0.39 to 0.96 for $MAX_e$ and 0.05 to 0.96 for $RMS_e$. Similarly, Compared with RSMC, the improvement ratios of SAC are ranging from 0.07 to 0.90 for $MAX_e$ and 0 to 0.91 for $RMS_e$. In addition, the presented SAC is considerably less sensitive to uncertainty than the other two controllers according to attitude tracking tests under disturbance. This has proven that the proposed SAC can improve performance robustness.

## 6 Conclusion

In this paper, the six degrees of freedom dynamic and kinematic equations of a tiltrotor UAV are derived. Based on the nonlinear mathematical model of tiltrotor UAV, a SAC is proposed, and the UAV simulations tests are established. The test results show that, compared with the FTSMC method and the RSMC method, the SAC method has higher control accuracy, faster response speed, and good robustness in tiltrotor UAV attitude control during transition mode (both conversion phase and reconversion phase). The SAC method can effectively control the roll angle, pitch angle, and yaw angle of the tiltrotor UAV.

Further researches about external disturbances and robustness analysis will be conducted to achieve trajectory tracking control. Besides, a full envelope flight test is planned to further study the flight control law.

## 7 Acknowledgments

This work was supported by the Science and Technology Planning Project of Guangdong under Grant 2017A020208063, and by the Science and Technology Planning Project of Guangzhou under Grant 201804010384.